\documentclass{article}
\usepackage{ijcai25}
\usepackage[table]{xcolor}
\usepackage{multirow}
\usepackage{times}
\usepackage{soul}
\usepackage{url}
\usepackage[hidelinks]{hyperref}
\usepackage[utf8]{inputenc}
\usepackage[small]{caption}
\usepackage{graphicx}
\usepackage{amsmath}
\usepackage{amsthm}
\usepackage{booktabs}
\usepackage{algorithm}
\usepackage{algorithmic}
\usepackage[switch]{lineno}
\usepackage{color}

\title{LDR-Net: A Novel Framework for AI-generated Image Detection via Localized Discrepancy Representation}

\author{
    JiaXin Chen, Miao Hu, DengYong Zhang, Yun Song,Xin Liao
}

\begin{document}

\maketitle

\begin{abstract}
With the rapid advancement of generative models, the visual quality of generated images has become nearly indistinguishable from the real ones, posing challenges to content authenticity verification. Existing methods for detecting AI-generated images primarily focus on specific forgery clues, which are often tailored to particular generative models like GANs or diffusion models. These approaches struggle to generalize across architectures. Building on the observation that generative images often exhibit local anomalies, such as excessive smoothness, blurred textures, and unnatural pixel variations in small regions, we propose the localized discrepancy representation network (LDR-Net), a novel approach for detecting AI-generated images. LDR-Net captures smoothing artifacts and texture irregularities, which are common but often overlooked. It integrates two complementary modules: local gradient autocorrelation (LGA) which models local smoothing anomalies to detect smoothing anomalies, and local variation pattern (LVP) which captures unnatural regularities by modeling the complexity of image patterns. By merging LGA and LVP features, a comprehensive representation of localized discrepancies can be provided. Extensive experiments demonstrate that our LDR-Net achieves state-of-the-art performance in detecting generated images and exhibits satisfactory generalization across unseen generative models. The code will be released upon acceptance of this paper.
\end{abstract}

\section{Introduction}
Recently, with the rapid development of generative models, such as Midjourney \cite{ruskov2023grimm} and DALL-E3 \cite{betker2023improving}, AI-generated images have achieved a visual quality that closely resembles real images, posing significant challenges to news communication and judicial authentication. This highlights the urgent need for automated detection methods that can accurately identify synthetic images and maintain the authenticity of visual content.
\cite{9897310} proposed an orthogonal training approach based on an ensemble of convolutional neural networks (CNNs), which aggregates features from multiple networks to detect synthetic images. However, its reliance on specific architectural patterns limits its performance when handling unseen generative models. \cite{sha2023fake} focused on detecting and attributing fake images generated by text-to-image models. Although high detection accuracy is achieved on diffusion models, their method shows limited adaptability to images generated by GANs. Most existing methods face challenges in addressing the diversity of generative models and the rapid evolution of generation techniques.

To address these limitations, we propose the Localized Discrepancy Representation Network (LDR-Net), motivated by the inherent constraints of current generative models, which often apply smoothing operations to ensure visual coherence. These operations result in excessive smoothness, blurred textures, and a lack of natural randomness in pixel intensity variations, which are key differences between real and generated images. Real images typically exhibit complex and diverse local patterns, intricate textures, and natural randomness in pixel distributions, while generated images often show uniformity, oversimplified details, and artifacts introduced during the generation process. LDR-Net introduces two complementary modules: the Local Gradient Autocorrelation (LGA) module, designed to detect smoothing anomalies in edge textures by modeling local gradient patterns, and the Local Variation Pattern (LVP) module, aimed at uncovering unnatural regularities in pixel intensity variations through directional encoding. These two modules comprehensively capture the discrepancies between real and generated images, providing a solid foundation for detection. Unlike existing methods that often rely heavily on generation-specific features, LDR-Net focuses on these generalized characteristics, enabling it to effectively generalize across diverse generative models, including unseen architectures and data distributions. The main contributions are as follows.

\begin{figure*}[htbp]
\centering
\includegraphics[width=6.2in]{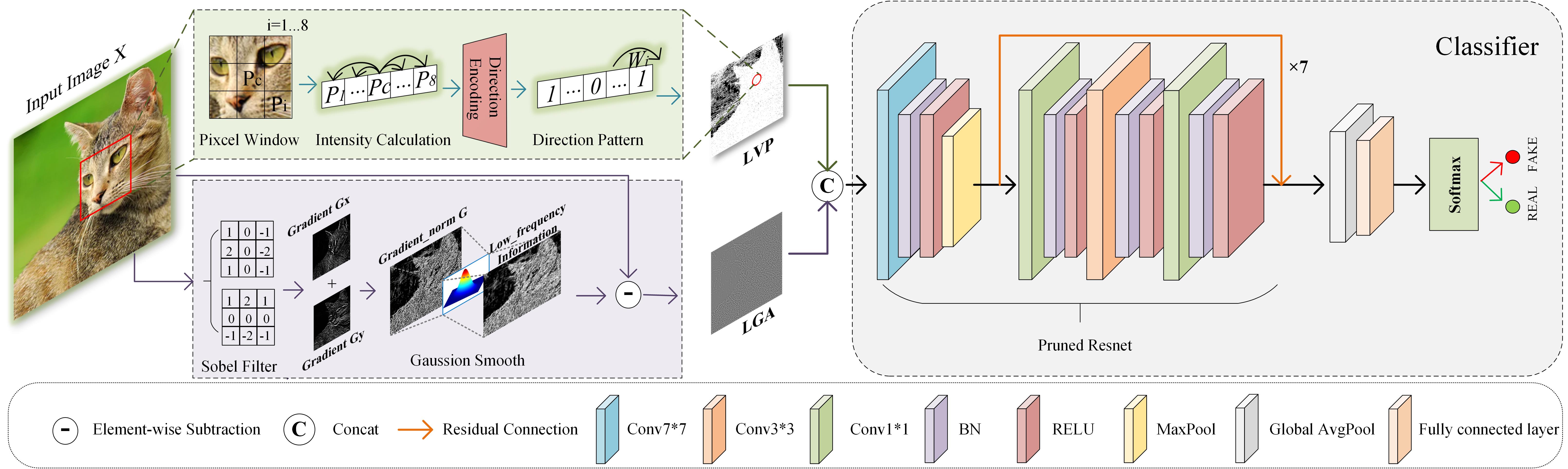}
\caption{Overview of the proposed LDR-Net framework for AI-generated image detection.}
\label{Figldr}
\vspace{-10pt}
\end{figure*}

\begin{itemize}
    \item We introduce local directional gradients and pixel coding patterns into AI-generated image detection, which brings a different viewpoint to cross-generator detection. We analyze the smoothing operation limitations during the generation of image generative models, and model the smoothing anomalies and unnatural pixel variations. Ultimately, the most differentiated gradient and pattern information can be mined.
    \item We propose a localized discrepancy representation network (LDR-Net) by integrating local gradient autocorrelation (LGA) and local variation pattern (LVP) to extract multi-level features. LGA detects high-frequency anomalies in edges and textures, while LVP captures low-frequency inconsistencies in pixel distributions. This fusion enables LDR-Net to robustly detect subtle discrepancies between real and generated images.
    \item We conduct extensive experiments to verify the effectiveness of LDR-Net in generated image detection, showcasing superior generalization ability on unseen generative models, including various GANs and diffusion models.
\end{itemize}

\section{Related Work}
\subsection{Local Feature-based Detection}
Local feature-based methods primarily focus on subtle inconsistencies in specific regions of generated images caused by the limitations of GANs in replicating natural patterns. For instance, \cite{liu2020global} proposed a method that enhances global texture to detect fake faces, which contrasts with our focus on localized anomalies in generated images. \cite{nguyen2019multi} introduced a forensic analysis method that detects interpolation artifacts in color filter arrays of digital images to uncover hidden anomalies. \cite{dong2022think} emphasized the role of spectral characteristics in generated images and proposed a frequency-domain analysis for distinguishing real and generated images. \cite{li2021detection} estimated the similarity of artificial artifacts in generated images, using their existence and distribution as key distinguishing features. Although these methods effectively detect specific types of artifacts, developing local feature extraction techniques with stronger robustness and generalization remains a significant challenge.

\subsection{Deep Learning-based Detection}
Deep learning-based methods have become mainstream for detecting AI-generated images due to their powerful feature extraction capabilities. \cite{marra2018detection} leveraged image content and contextual information to improve detection on social networks. While \cite{wang2020cnn} utilized adversarial training and feature matching to enhance accuracy. To tackle cross-domain challenges, \cite{tan2023learning} learned gradient information to capture subtle structural differences, boosting generalization across generators. \cite{zhang2022improving} introduced unsupervised domain adaptation to adapt models to unseen data. Additionally, \cite{lim2024distildire} designed a lightweight diffusion synthesis detector to reduce computational demands, and \cite{safwat2024hybrid} proposed a hybrid GAN-ResNet model for robust fake face detection.

Different from the above approaches, our LDR-Net focuses on improving generalization to unseen manipulations by refining local feature extraction. By capturing anomalies in texture consistency and pixel distribution patterns, it effectively highlights discrepancies between generated and real images, achieving strong robustness and adaptability even against unknown generative techniques.

\section{Proposed Method}
As shown in Fig. \ref{Figldr}, this paper proposes the localized discrepancy representation network (LDR-Net), which is composed of local gradient autocorrelation (LGA), local variation pattern (LVP), and a classifier. The LGA features capture local smoothing anomalies through autocorrelation calculations of local gradients, while the LVP features detect anomalies in the lack of complex variation patterns via directional encoding between pixels. These two types of features reveal the differences between generated and real images from the perspectives of edge texture and pixel distribution. The extracted LGA and LVP features are concatenated and fed into the classifier to complete the AI-generated image detection task. The detailed design is introduced in this section.

\subsection{Local Gradient Autocorrelation}
Real images typically exhibit high texture consistency and complex detail patterns in their local regions. However, when generating synthetic images, generators often introduce smoothing operations to ensure overall visual quality. This results in excessive smoothness or texture blurring in the local regions of generated images. In this case, we design a local gradient autocorrelation module (LGA) to describe the inevitable local smoothness anomalies in generated images.

\begin{figure}[tbp]
\centering
\includegraphics[scale=0.4]{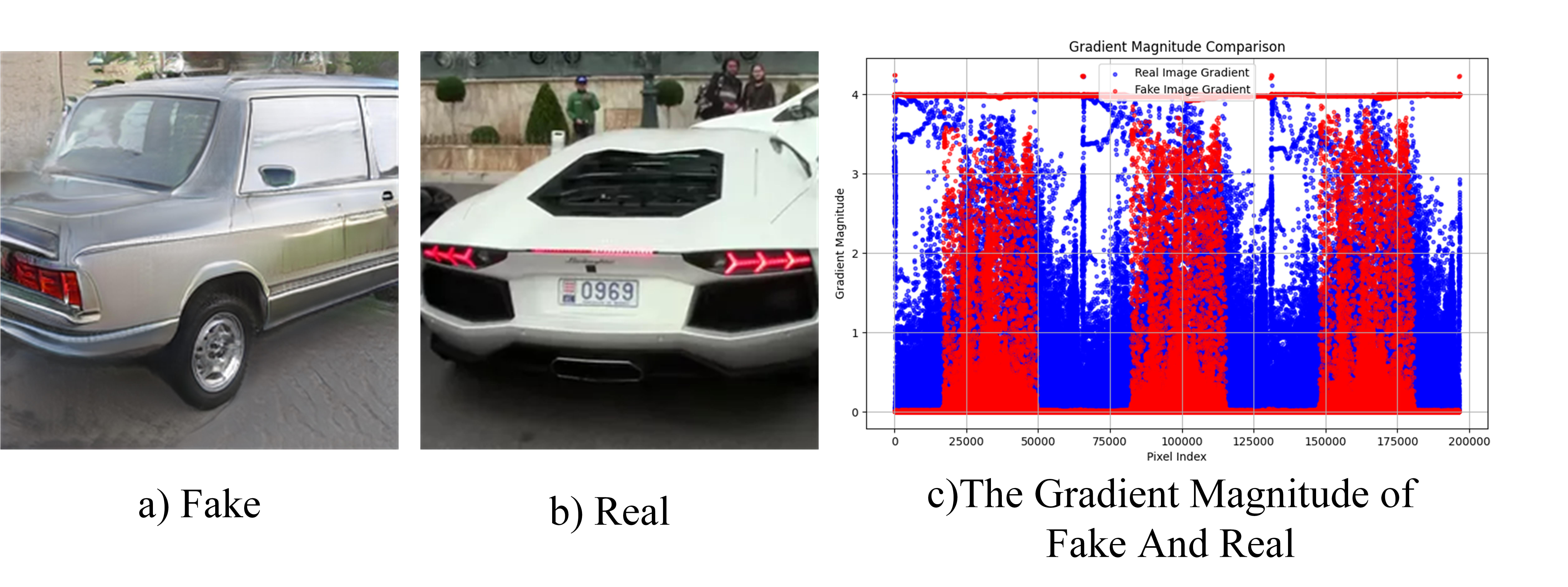}
\caption{Illustration of gradient differences between generated images and real images. a) is a fake image generated from the real image b). c) is a scatter plot that statistically represents the gradient magnitude changes between corresponding pixels in a) and b). Our LDR-Net effectively reveals the variation of gradient magnitude between generated images and real images.}
\label{Fig1}
\end{figure}

Unlike traditional methods that extract gradient directions and compute global gradient features in a block-wise manner, our approach focuses exclusively on the variation of local gradient features. The primary distinction between the generated image and the real image lies not in global differences but in the similarity of adjacent pixel patterns within local regions. It can be deduced that the local areas of the generated image exhibit significant deviations from those of the real image. Focusing on changes in local directional gradients proves to be more effective for detecting forged traces. Fig. \ref{Fig1} computes the gradient magnitude variations between pixels in real and generated images, revealing significant differences between them, which can be served as discriminative features for distinguishing generated images from real ones.

To further capture edge and texture information, we utilize the Sobel operator to extract local gradient features from the images. For an input image $X(B \times C \times H \times W)$, where $B$ is the batch size, $C$ is the number of channels, and $H$ and $W$ are the height and width of the image. Sobel convolution kernels $W_x$ and $W_y$ in the horizontal gradient and vertical gradient are defined as follows.
\begin{equation}
	\begin{gathered}
	 W_{x}=\begin{bmatrix}1 & 0 & -1 \\2 & 0 & -2 \\1 & 0 & -1\end{bmatrix}	
	\end{gathered}
    \begin{gathered}
    ,W_{y}=\begin{bmatrix}1 & 2 & 1 \\0 & 0 & 0 \\-1 & -2 & -1\end{bmatrix}	
	\end{gathered}
\end{equation}

We first perform convolution on the image $X$ to obtain the horizontal gradient $G_x$ and vertical gradient $G_y$.
\begin{equation}
	G_{x}=Conv(W_{x}, X), G_{Y}=Conv(W_{y}, X)
\end{equation}

The gradient magnitude $G$ is then computed from $G_x$ and $G_y$ to capture the local edge and detail features of the image, 
\begin{equation}
    G=\sqrt{G_{x}^2+G_{y}^2+\epsilon}
\end{equation}
where $\epsilon$ is a constant introduced to avoid zero-value issues in gradient magnitude computation. 

Generators frequently encounter challenges in synthesizing high-frequency details, such as textures and edges, and may attempt to obscure forgeries by adding noise. Gaussian smoothing can effectively mitigate high-frequency noise in gradient features, thereby accentuating anomalies within the generated image, which is crucial for detecting generated images. The gradient feature $G$ focuses on edge and texture information, which are typically the most distinguishable areas between generated and real images. If Gaussian smoothing is  applied directly to the input image $X$, edge and texture information may be blurred, leading to weakened key features. By first computing the gradient feature $G$, high-frequency details can be preserved, enhancing the detection of local anomalies. Therefore, we apply Gaussian smoothing convolution to the gradient feature $G$ to reduce the smoothness and consistency of local regions. Specifically, a Gaussian kernel $K(x, y)$ is used to model the smoothing variations of local regions,
\begin{equation}
    K(x, y) = \frac{1}{2\pi\sigma^2} e^{-\frac{x^2 + y^2}{2\sigma^2}}
    \label{eq:gaussian_kernel}
\end{equation}
where $x$ and $y$ represent the pixel offsets relative to the kernel center, and $\sigma$ is a hyperparameter that determines the strength of the smoothing effect. A larger $\sigma$ results in stronger smoothing, while a smaller $\sigma$ preserves more local details.

We perform convolution of the feature $G$ with the Gaussian kernel $K(x, y)$ to obtain the autocorrelation feature $A$.
\begin{equation}
    A = Conv(K(x, y), G)
\end{equation}

Feature $A$ is smoothed to suppress the high-frequency noise typically introduced during the generation process and preserve low-frequency information. By computing the residual between $G$ and $A$, the high-frequency components are effectively highlighted, while the impact of less relevant features is minimized. This process yields the local gradient autocorrelation (LGA) feature.
\begin{equation}
    \text{LGA} = G - A
\end{equation}

The LGA feature effectively emphasizes high-frequency anomalies and smoothing characteristics in local regions, providing critical evidence for detecting generated images.

\subsection{Local Variation Pattern}
Local regions in real images typically exhibit complex pixel distributions and diverse variation patterns. However, due to the inherent limitations of image generators, the pixel distribution of generated images tends to be more uniform. Inspired by this observation, we propose a local variation pattern (LVP) module based on pixel intensity relationships, which aims to describe the relative variations between local pixels and reveal potential anomalies in generated images.

For each pixel $P_c$ in the input image $X(B \times C \times H \times W)$, we define a $3 \times 3$ local neighborhood window centered on $P_c$, which includes the pixel itself and its 8 surrounding neighboring pixels. To capture the local variation, we compute the intensity differences between the central pixel $P_c$ and each of its neighboring pixels. The calculation can be expressed as,
\begin{equation}
    \Delta I(P_c, P_n) = I(P_c) - I(P_n), \quad P_n \in \mathcal{N}(P_c)
\end{equation}
where $I(P_c)$ and $I(P_n)$ represent the intensities of the central pixel $P_c$ and a neighboring pixel $P_n$, respectively. $\mathcal{N}(P_c)$ denotes the set of all eight neighboring pixels around $P_c$. $\Delta I(P_c, P_n)$ represents the intensity difference between the central pixel and its neighboring pixels. Each difference value $\Delta I(P_c, P_n)$ is converted into a symbolic directional code to represent the variation trend of the neighboring pixel intensity relative to the central pixel. 

\begin{figure}[tbp]
\centering
\includegraphics[width=\columnwidth]{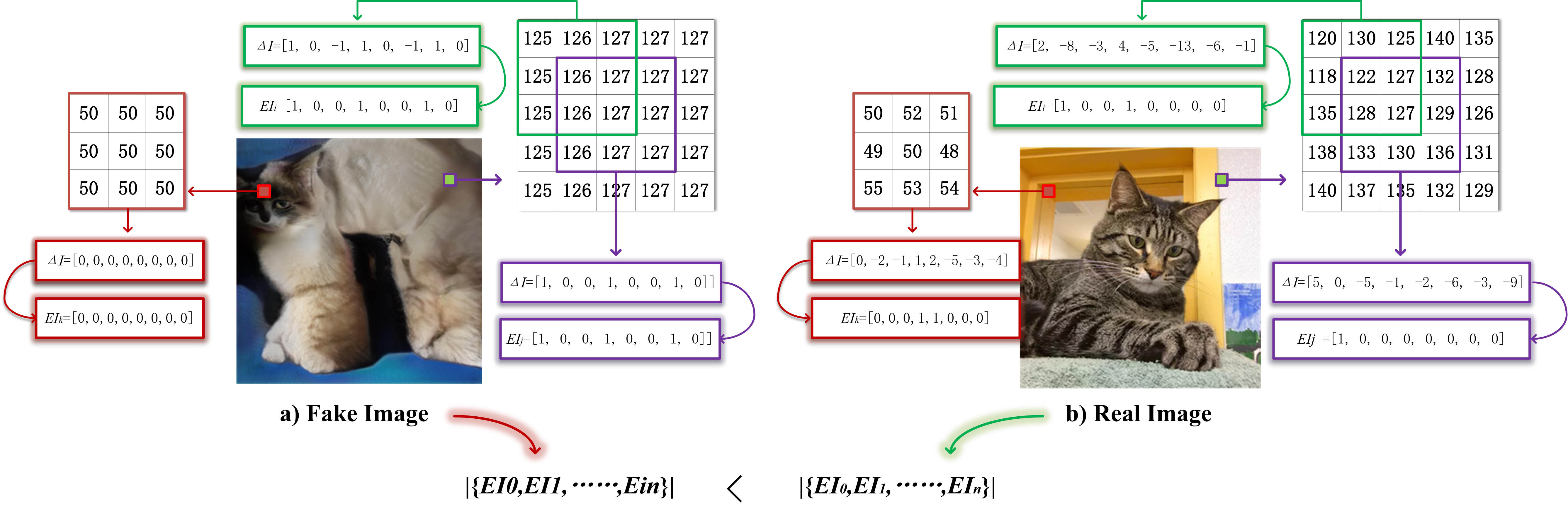}
\caption{Comparison of local binary encoding patterns between real and generated images, a) is fake image generated from real image b).}
\label{Fig2}
\end{figure}

This abstraction transforms the relationship between the central pixel and its neighboring pixels into a binary pattern, inherently reflecting the complexity and diversity of the local region. As illustrated in Fig. \ref{Fig2}, due to the inherent limitations of the generator, local regions of the generated image lack the randomness observed in real scenarios, leading to highly similar directional encoding. In extreme cases, when large local regions in a generated image consist of identical pixel values, the directional encoding may exhibit fixed patterns, such as all zeros or all ones. Conversely, real images show highly complex textures and edges, producing more diverse binary patterns in directional encoding. As a result, the pattern complexity of generated images is significantly lower compared to that of real images. Let $E_i(p)$ denote the directional encoding of the $i$th neighboring pixel relative to pixel $p$, which maps the complexity of these patterns.
\begin{equation}
E_i(p) =
\begin{cases}
1, & \Delta I(P_c, P_n) > 0, \\
0, & \Delta I(P_c, P_n) \leq 0
\end{cases}
\end{equation}

To generate the local variation pattern feature, we assign a unique weight to each direction and perform a weighted aggregation of the directional encoding for pixel $p$. To ensure the uniqueness of the  feature and avoid confusion in the directional encoding results, we randomly select eight distinct real numbers $W_i$ ($i = 0, 1, \dots, 7$) as weights. This weight design ensures that each direction has a unique and non-repeating weight, and that there is no ambiguity between adjacent directions. Consequently, the unique feature can preserve the variation information of each direction completely.
\begin{equation}
LVP(p)=\sum_{i=0}^{N} E_i(p)\times W_i (N=7)
\end{equation}

In real images, directional variation patterns are typically highly diverse. The relative intensity relationships in different directions form a wide range of combinations, resulting in a broad distribution of feature values. While in generated images, variation patterns across multiple directions tend to be uniform, leading to a loss of directional information and producing feature values with a narrow or overly simplistic distribution. Thus, the local variation pattern feature value, designed with unique weights, effectively reveals anomalous pattern differences between real and generated images.

By repeating the above steps for each pixel $p$ in the input image $X$, the LVP feature can be extracted, which effectively encodes the variation patterns within local regions of the image, providing a reliable basis for analyzing the differences between real and generated images.

\subsection{Classifier}
Our classifier is based on a pruned ResNet \cite{he2016deep} as the backbone, where structural optimizations significantly reduce network parameters and computational complexity while preserving efficient feature extraction capabilities. During the feature processing stage, LVP and LGA features are integrated through feature-level concatenation (concat), capturing both local variation patterns and global feature correlations. Subsequently, a Global Average Pooling (GAP) layer is employed to compress spatial information effectively, generating a compact global feature representation. Finally, the fully connected layer maps the high-dimensional features into a binary classification space to predict whether the input image is generated or real.

\begin{table*}[]
\centering
\caption{Ablation studies of several modules on the ForenSynths datasets.}
\renewcommand\arraystretch{1.1}
\resizebox{\textwidth}{!}{ 
\begin{tabular}{ccccccccccccccccc|cc}
\hline
                                         & \multicolumn{2}{c}{\textbf{ProGAN}} & \multicolumn{2}{c}{\textbf{StyleGAN}}       & \multicolumn{2}{c}{\textbf{StyleGAN2}}      & \multicolumn{2}{c}{\textbf{BigGAN}}        & \multicolumn{2}{c}{\textbf{CycleGAN}} & \multicolumn{2}{c}{\textbf{StarGAN}}      & \multicolumn{2}{c}{\textbf{GauGAN}}   & \multicolumn{2}{c|}{\textbf{Deepfake}}      & \multicolumn{2}{c}{\textbf{Mean}} \\ \cline{2-19} 
\multirow{-2}{*}{\textbf{Method}}        & \textbf{ACC}      & \textbf{AP}     & \textbf{ACC}          & \textbf{AP}         & \textbf{ACC}          & \textbf{AP}         & \textbf{ACC}          & \textbf{AP}        & \textbf{ACC}       & \textbf{AP}      & \textbf{ACC}         & \textbf{AP}        & \textbf{ACC}       & \textbf{AP}      & \textbf{ACC}          & \textbf{AP}         & \textbf{ACC}    & \textbf{AP}     \\ \hline
\textbf{LGA}                             & 99.8              & 100.0           & 94.8                  & 99.9                & 96.3                  & 99.9                & 81.5                  & 89.3               & 91.5               & 97.1             & 92.4                 & 99.9               & 74.8               & 77.0             & 51.9                  & 76.5                & 85.4            & 92.4            \\
\textbf{LVP}                             & 99.9              & 100.0           & 98.1                  & 99.8                & 95.9                  & 99.3                & 83.4                  & 83.3               & 79.3               & 82.4             & 99.4                 & 100.0              & 71.3               & 77.2             & 59.2                  & 77.3                & 85.8            & 89.9            \\ 
\textbf{LDR-Net (Roberts)}                             & 96.4              & 99.5           & 90.6                  & 97.7                & 90.9                  & 98.1                & 88.8                  & 94.7               & 70.2               & 83.2             & 97.8                 & 100.0              & 68.5               & 70.8             & 53.2                  & 64.5                & 82.1            & 88.6            \\ 
\textbf{LDR-Net (Canny)}                             & 99.8              & 100.0           & 98.1                  & 99.9                & 95.9                  & 99.8                & 91.0                  & 96.6               & 90.7               & 96.5             & 99.6                 & 100.0              & 77.8               & 79.7             & 56.6                  & 74.0                & 88.7            & 93.3            \\ 
c
\textbf{LDR-Net (Sobel)}             & 99.9         & 100.0                & 97.9              & 99.9                    & 96.5               & 99.8                   & 91.4             & 97.1                    & 91.4           & 97.6                 & 99.7                  & 100.0                & 80.8           & 86.7                 & 68.6          & 81.5                                              & \textbf{90.8}            & \textbf{95.3}       \\ \hline
                                        
\end{tabular}
}
\label{tab:ablation}
\end{table*}
\section{Experiments}

\begin{table*}[]
\centering
\caption{Cross-GAN sources evaluation on the ForenSynths dataset. The best performance is highlighted using bold and underlined text, while the second-best performance is highlighted using bold text.}

\renewcommand\arraystretch{1.2}
\resizebox{\textwidth}{!}{ 
\begin{tabular}{ccccccccccccccccc|cc}
\hline
                                  & \multicolumn{2}{c}{\textbf{ProGAN}} & \multicolumn{2}{c}{\textbf{StyleGAN}} & \multicolumn{2}{c}{\textbf{StyleGAN2}} & \multicolumn{2}{c}{\textbf{BigGAN}} & \multicolumn{2}{c}{\textbf{CycleGAN}} & \multicolumn{2}{c}{\textbf{StarGAN}} & \multicolumn{2}{c}{\textbf{GauGAN}} & \multicolumn{2}{c|}{\textbf{DeepFake}} & \multicolumn{2}{c}{\textbf{Mean}} \\ \cline{2-19} 
\multirow{-2}{*}{\textbf{Method}} & \textbf{ACC}      & \textbf{AP}     & \textbf{ACC}       & \textbf{AP}      & \textbf{ACC}       & \textbf{AP}       & \textbf{ACC}      & \textbf{AP}     & \textbf{ACC}       & \textbf{AP}      & \textbf{ACC}      & \textbf{AP}      & \textbf{ACC}      & \textbf{AP}     & \textbf{ACC}       & \textbf{AP}       & \textbf{ACC}    & \textbf{AP}     \\ \hline
\textbf{CNNDetection (CVPR'20)}             & 91.4              & 99.4            & 63.8               & 91.4             & 76.4               & 97.5              & 52.9              & 73.3            & 72.7               & 88.6             & 63.8              & 90.8             & 63.9              & 92.2            & 51.7               & 62.3              & 67.1            & 86.9            \\
\textbf{Frank (PRML'20)}                    & 90.3              & 85.2            & 74.5               & 72.0             & 73.1               & 71.4              & 88.7              & 86.0            & 75.5               & 71.2             & 99.5              & 99.5             & 60.2              & 77.4            & 60.7               & 49.1              & 78.9            & 76.5            \\
\textbf{Durall (CVPR'20)}                   & 81.1              & 74.4            & 54.4               & 52.6             & 66.8               & 62.0              & 60.1              & 56.3            & 69.0               & 64.0             & 98.1              & 98.1             & 61.9              & 57.4            & 50.2               & 50.0              & 67.7            & 64.4            \\
\textbf{Patchfor (ECCV'20)}                 & 97.8              & 100.0           & 82.6               & 93.1             & 83.6               & 98.5              & 64.7              & 69.5            & 74.5               & 87.2             & 100.0             & 100.0            & 57.2              & 55.4            & 85.0               & 93.2              & 80.7            & 87.1            \\
\textbf{F3Net (ECCV'20)}                    & 99.4              & 100.0           & 92.6               & 99.7             & 88.0               & 99.8              & 65.3              & 69.9            & 76.4               & 84.3             & 100.0             & 100.0            & 58.1              & 56.7            & 63.5               & 78.8              & 80.4            & 86.2            \\
\textbf{SelfBland (CVPR'22)}                & 58.8              & 65.2            & 50.1               & 47.7             & 48.6               & 47.4              & 51.1              & 51.9            & 59.2               & 65.3             & 74.5              & 89.2             & 59.2              & 65.5            & 93.8               & 99.3              & 61.9            & 66.4            \\
\textbf{GANDetection (ICIP'22)}             & 82.7              & 95.1            & 74.4               & 92.9             & 69.9               & 87.9              & 76.3              & 89.9            & 85.2               & 95.5             & 68.8              & 99.7             & 61.4              & 75.8            & 60.0               & 83.9              & 72.3            & 90.1            \\
\textbf{BiHPF (WACV'22)}                    & 90.7              & 86.2            & 76.9               & 75.1             & 76.2               & 74.7              & 84.9              & 81.7            & 81.9               & 78.9             & 94.4              & 94.4             & 69.5              & 78.1            & 54.4               & 54.6              & 78.6            & 77.9            \\
\textbf{FrePGAN (AAAI'22)}                  & 99.0              & 99.9            & 80.7               & 89.6             & 84.1               & 98.6              & 69.2              & 71.1            & 71.1               & 74.4             & 99.9              & 100.0            & 60.3              & 71.7            & 70.9               & 91.9              & 79.4            & 87.2            \\
\textbf{LGrad (CVPR'23)}                    & 99.9              & 100.0           & 94.8               & 99.9             & 96.0               & 99.9              & 82.9              & 90.7            & 85.3               & 94.0             & 99.6              & 100.0            & 72.4              & 79.3            & 58.0               & 67.9              & 86.1            & 91.5            \\
\textbf{Ojha (CVPR'23)}                     & 99.7              & 100.0           & 89.0               & 98.7             & 83.9               & 98.4              & 90.5              & 99.1            & 87.9               & 99.8             & 91.4              & 100.0            & 89.9              & 100.0           & 80.2               & 90.2              & 89.1            & \textbf{\underline{98.3}}   \\
\textbf{MI\_Net (AAAI'24)}                    & 99.2             & 100.0           & 89.1               & 96.1             & 96.9               & 99.7              & 62.8              & 60.6         & 68.6               & 74.5             & 99.7              & 100.0            & 54.7              & 50.6            & 73.5              & 81.5              & 80.6            & 82.9            \\
\textbf{NPR (CVPR'24)}                     & 99.9              & 100.0           & 96.4               & 99.9             & 97.0               & 99.9              & 85.3              & 91.4            & 85.6               & 98.7             & 99.8              & 100.0            & 83.0              & 84.1            & 79.6               & 85.4              & \textbf{{\underline{90.8}}}   & 94.9            \\
\rowcolor[HTML]{C0C0C0} 
\textbf{LDR-Net (our)}             & 99.9         & 100.0                & 97.9              & 99.9                    & 96.5               & 99.8                   & 91.4             & 97.1                    & 91.4           & 97.6                 & 99.7                  & 100.0                & 80.8           & 86.7                 & 68.6          & 81.5                                              & \textbf{\underline{90.8}}            & \textbf{95.3}            \\
 \hline
\end{tabular}
}
\label{tab:cross_gan_evaluation_1}
\end{table*}

\begin{table*}[]
\centering
\caption{Cross-Diffusion sources evaluation on the DiffusionForensics dataset. The best performance is highlighted using bold and underlined text, while the second-best performance is highlighted using bold text.}
\renewcommand\arraystretch{1.2}

\resizebox{\textwidth}{!}{ 
\begin{tabular}{ccccccccccccccccc|cc}
\hline
                                  & \multicolumn{2}{c}{\textbf{ADM}} & \multicolumn{2}{c}{\textbf{DDPM}} & \multicolumn{2}{c}{\textbf{IDDPM}} & \multicolumn{2}{c}{\textbf{LDM}} & \multicolumn{2}{c}{\textbf{PNDM}} & \multicolumn{2}{c}{\textbf{VQ-Diffusion}} & \multicolumn{2}{c}{\textbf{Sdv1}} & \multicolumn{2}{c|}{\textbf{Sdv2}} & \multicolumn{2}{c}{\textbf{Mean}} \\ \cline{2-19} 
\multirow{-2}{*}{\textbf{Method}} & \textbf{ACC}    & \textbf{AP}    & \textbf{ACC}     & \textbf{AP}    & \textbf{ACC}     & \textbf{AP}     & \textbf{ACC}    & \textbf{AP}    & \textbf{ACC}     & \textbf{AP}    & \textbf{ACC}         & \textbf{AP}        & \textbf{ACC}     & \textbf{AP}    & \textbf{ACC}     & \textbf{AP}     & \textbf{ACC}    & \textbf{AP}     \\ \hline
\textbf{CNNDetection (CVPR'20)}             & 53.9            & 71.8           & 62.7             & 76.6           & 50.2             & 82.7            & 50.4            & 78.7           & 50.8             & 90.3           & 50.0                 & 71.0               & 38.0             & 76.7           & 52.0             & 90.3            & 51.0            & 79.8            \\
\textbf{Frank (PRML'20)}                    & 58.9            & 65.9           & 37.0             & 27.6           & 51.4             & 65.0            & 51.7            & 48.5           & 44.0             & 38.2           & 51.7                 & 66.7               & 32.8             & 52.3           & 40.8             & 37.5            & 46.0            & 50.2            \\
\textbf{Durall (CVPR'20)}                   & 39.8            & 42.1           & 52.9             & 49.8           & 55.3             & 56.7            & 43.1            & 39.9           & 44.5             & 47.3           & 38.6                 & 38.3               & 39.5             & 56.3           & 62.1             & 55.8            & 47.0            & 48.3            \\
\textbf{Patchfor (ECCV'20)}                 & 77.5            & 93.9           & 62.3             & 97.1           & 50.0             & 91.6            & 99.5            & 100.0          & 50.2             & 99.9           & 100.0                & 100.0              & 90.7             & 99.8           & 94.8             & 100.0           & 78.1            & 97.8            \\
\textbf{F3Net (ECCV'20)}                    & 80.9            & 96.9           & 84.7             & 99.4           & 74.7             & 98.9            & 100.0           & 100.0          & 72.8             & 99,5           & 100.0                & 100.0              & 73.4             & 97.2           & 99.8             & 100.0           & 85.8            & \textbf{99.0}            \\
\textbf{SelfBland (CVPR'22)}                & 57.0            & 59.0           & 61.9             & 49.6           & 63.2             & 66.9            & 83.3            & 92.2           & 48.2             & 48.2           & 77.2                 & 82.7               & 46.2             & 68.0           & 71.2             & 73.9            & 63.5            & 67.6            \\
\textbf{GANDetection (ICIP'22)}             & 51.1            & 53.1           & 62.3             & 46.4           & 50.2             & 63.0            & 51.6            & 48.1           & 50.6             & 79.0           & 51.1                 & 51.2               & 39.8             & 65.6           & 50.1             & 36.9            & 50.8            & 55.4            \\
\textbf{LGrad (CVPR'23)}                    & 86.4            & 97.5           & 99.9             & 100.0          & 66.1             & 92.8            & 99.7            & 100.0          & 69.5             & 98.5           & 96.2                 & 100.0              & 90.4             & 99.4           & 97.1             & 100.0           & 88.2            & 98.5            \\
\textbf{Ojha (CVPR'23)}                     & 78.4            & 92.1           & 72.9             & 78.8           & 75.0             & 92.8            & 82.2            & 97.1           & 75.3             & 92.5           & 83.5                 & 97.7               & 56.4             & 90.4           & 71.5             & 92.4            & 74.4            & 91.7            \\
\textbf{MI\_Net (AAAI'24)}                    & 83.3             & 90.0           & 62.5               & 54.6             & 50.0              & 55.7              & 99.3              & 100.0         & 56.2               &  99.1             & 99.7              & 100.0            & 95.7              & 99.6           & 99.2              & 100.0              & 80.7          & 87.4            \\

\textbf{NPR (CVPR'24)}                     & 86.3            & 98.0           & 90.7             & 98.2           & 84.3             & 92.2            & 100.0           & 100.0          & 93.7             & 100.0          & 96.6                & 99.9              & 98.9             & 100.0           & 100.0             & 100.0           & \textbf{93.8}            & 98.5            \\
\rowcolor[HTML]{C0C0C0} 
\textbf{LDR-Net (our)}                  & 92.5            & 98.5           & 98.8             & 100.0          & 97.0             & 99.7            & 99.1            & 100.0          & 98.9             & 100.0          & 99.1                 & 100.0              & 85.1             & 98.5           & 97.6             & 99.8            & \textbf{\underline{96.0}}   & \textbf{\underline{99.6}}   \\ \hline
\end{tabular}

}
\label{tab:cross_gan_evaluation_3}
\end{table*}

\begin{table*}[]
\centering
\caption{Cross-Diffusion sources evaluation on the Ojha dataset. The best performance is highlighted using bold and underlined text, while the second-best performance is highlighted using bold text.}
\renewcommand\arraystretch{1.2}

\resizebox{\textwidth}{!}{ 
\begin{tabular}{ccccccccccccccccc|cc}
\hline
                                  & \multicolumn{2}{c}{\textbf{DALLE}} & \multicolumn{2}{c}{\textbf{Glide\_100\_10}} & \multicolumn{2}{c}{\textbf{Glide\_100\_27}} & \multicolumn{2}{c}{\textbf{Glide\_50\_27}} & \multicolumn{2}{c}{\textbf{ADM}} & \multicolumn{2}{c}{\textbf{LDM\_100}} & \multicolumn{2}{c}{\textbf{LDM\_200}} & \multicolumn{2}{c|}{\textbf{LDM\_200\_cfg}} & \multicolumn{2}{c}{\textbf{Mean}} \\ \cline{2-19} 
\multirow{-2}{*}{\textbf{Method}} & \textbf{ACC}     & \textbf{AP}     & \textbf{ACC}          & \textbf{AP}         & \textbf{ACC}          & \textbf{AP}         & \textbf{ACC}         & \textbf{AP}         & \textbf{ACC}    & \textbf{AP}    & \textbf{ACC}       & \textbf{AP}      & \textbf{ACC}       & \textbf{AP}      & \textbf{ACC}          & \textbf{AP}         & \textbf{ACC}    & \textbf{AP}     \\ \hline
\textbf{CNNDetection (CVPR'20)}             & 51.8             & 61.3            & 53.3                  & 72.9                & 53.0                  & 71.3                & 54.2                 & 76.0                & 54.9            & 66.6           & 51.9               & 63.7             & 52.0               & 64.5             & 51.6                  & 63.1                & 52.8            & 67.4            \\
\textbf{Frank (PRML'20)}                    & 57.0             & 62.5            & 53.6                  & 44.3                & 50.4                  & 40.8                & 52.0                 & 42.3                & 53.4            & 52.5           & 56.6               & 51.3             & 56.4               & 50.9             & 56.5                  & 52.1                & 54.5            & 49.6            \\
\textbf{Durall (CVPR'20)}                   & 55.9             & 58.0            & 54.9                  & 52.3                & 48.9                  & 46.9                & 51.7                 & 49.9                & 40.6            & 42.3           & 62.0               & 62.6             & 61.7               & 61.7             & 58.4                  & 58.5                & 54.3            & 54.0            \\
\textbf{Patchfor (ECCV'20)}                 & 79.8             & 99.1            & 87.3                  & 99.7                & 82.8                  & 99.1                & 84.9                 & 98.8                & 74.2            & 81.4           & 95.8               & 99.8             & 95.6               & 99.9             & 94.0                  & 99.8                & 86.8            & 97.2            \\
\textbf{F3Net (ECCV'20)}                    & 71.6             & 79.9            & 88.3                  & 95.4                & 87.0                  & 94.5                & 88.5                 & 95.4                & 69.2            & 70.8           & 74.1               & 84.0             & 73.4               & 83.3             & 80.7                  & 89.1                & 79.1            & 86.5            \\
\textbf{SelfBland (CVPR'22)}                & 52.4             & 51.6            & 58.8                  & 63.2                & 59.4                  & 64.1                & 64.2                 & 68.3                & 58.3            & 63.4           & 53.0               & 54.0             & 52.6               & 51.9             & 51.9                  & 52.6                & 56.3            & 58.7            \\
\textbf{GANDetection (ICIP'22)}             & 67.2             & 83.0            & 51.2                  & 52.6                & 51.1                  & 51.9                & 51.7                 & 53.5                & 49.6            & 49.0           & 54.7               & 65.8             & 54.9               & 65.9             & 53.8                  & 58.9                & 54.3            & 60.1            \\
\textbf{LGrad (CVPR'23)}                    & 88.5             & 97.3            & 89.4                  & 94.9                & 87.4                  & 93.2                & 90.7                 & 95.1                & 86.6            & 100.0          & 94.8               & 99.2             & 94.2               & 99.1             & 95.9                  & 99.2                & \textbf{90.9}            & 97.2            \\
\textbf{Ojha (CVPR'23)}                     & 89.5             & 96.8            & 90.1                  & 97.0                & 90.7                  & 97.2                & 91.1                 & 97.4                & 75.7            & 85.1           & 90.5               & 97.0             & 90.2               & 97.1             & 77.3                  & 88.6                & 86.9            & 94.5            \\
\textbf{MI\_Net (AAAI'24)}                    & 63.3             & 73.6           & 80.8               & 90.0            & 78.5               & 88.0              & 82.0              & 90.3         & 69.5               & 75.5             & 81.3              & 90.5           & 81.2             & 89.9           & 80.0              & 89.3              & 77.1            & 85.9           \\
\textbf{NPR (CVPR'24)}                     & 75.6             & 98.3            & 96.6                  & 99.8                & 95.8                  & 99.8                & 99.8                 & 99.6                & 76.0            & 87.5           & 95.4               & 99.9             & 95.2               & 99.9             & 95.5                  & 99.8                & 90.8            & \textbf{98.1}            \\
\rowcolor[HTML]{C0C0C0} 
\textbf{LDR-Net (our)}                  & 93.2             & 98.9            & 95.0                  & 99.1                & 90.0                  & 97.3                & 92.2                 & 97.8                & 87.6            & 94.7           & 95.8               & 99.3             & 94.9               & 99.1             & 95.9                  & 99.3                & \textbf{\underline{93.1}}   & \textbf{\underline{98.2}}   \\ \hline
\end{tabular}
}
\label{tab:cross_gan_evaluation_4}
\end{table*}

\begin{table*}[]
\centering
\caption{Cross-Diffusion sources evaluation on the Self-Synthesis dataset. The best performance is highlighted using bold and underlined text, while the second-best performance is highlighted using bold text.}
\renewcommand\arraystretch{0.96}
\resizebox{\textwidth}{!}{ 
\tiny
\begin{tabular}{ccccccccccc|cc}
\hline
                                  & \multicolumn{2}{c}{\textbf{DDPM}} & \multicolumn{2}{c}{\textbf{IDDPM}} & \multicolumn{2}{c}{\textbf{ADM}} & \multicolumn{2}{c}{\textbf{Midjourney}} & \multicolumn{2}{c|}{\textbf{DALLE}} & \multicolumn{2}{c}{\textbf{Mean}} \\ \cline{2-13} 
\multirow{-2}{*}{\textbf{Method}} & \textbf{ACC}     & \textbf{AP}    & \textbf{ACC}     & \textbf{AP}     & \textbf{ACC}    & \textbf{AP}    & \textbf{ACC}        & \textbf{AP}       & \textbf{ACC}      & \textbf{AP}     & \textbf{ACC}    & \textbf{AP}     \\ \hline
\textbf{CNNDetection (CVPR'20)}             & 50.0             & 63.3           & 48.3             & 52.7            & 53.4            & 64.4           & 48.6                & 38.5              & 49.3              & 44.7            & 49.9            & 52.7            \\
\textbf{Frank (PRML'20)}                    & 47.6             & 43.1           & 70.5             & 85.7            & 67.3            & 72.2           & 39.7                & 40.8              & 68.7              & 65.2            & 58.8            & 61.4            \\
\textbf{Durall (CVPR'20)}                   & 54.1             & 53.6           & 63.2             & 71.7            & 39.1            & 40.8           & 45.7                & 47.2              & 53.9              & 52.2            & 51.2            & 53.1            \\
\textbf{Patchfor (ECCV'20)}                 & 54.1             & 66.3           & 35.8             & 34.2            & 68.6            & 73.7           & 66.3                & 68.8              & 60.8              & 65.1            & 57.1            & 61.6            \\
\textbf{F3Net (ECCV'20)}                    & 59.4             & 71.9           & 42.2             & 44.7            & 73.4            & 80.3           & 73.2                & 80.4              & 79.6              & 87.3            & 65.5            & 72.9            \\
\textbf{SelfBland (CVPR'22)}                & 55.3             & 57.7           & 63.5             & 62.5            & 57.1            & 60.1           & 54.3                & 56.4              & 48.8              & 47.4            & 55.8            & 56.8            \\
\textbf{GANDetection (ICIP'22)}             & 47.3             & 45.5           & 47.9             & 57.0            & 51.0            & 56.1           & 50.0                & 44.7              & 49.8              & 49.7            & 49.2            & 50.6            \\
\textbf{LGrad (CVPR'23)}                    & 59.8             & 88.5           & 45.2             & 46.9            & 72.7            & 79.3           & 68.3                & 76.0              & 75.1              & 80.9            & 64.2            & 74.3            \\
\textbf{Ojha (CVPR'23)}                     & 69.5             & 80.0           & 64.9             & 74.2            & 81.3            & 90.8           & 50.0                & 49.8              & 66.3              & 74.6            & 66.4            & 73.9            \\
\textbf{MI\_Net (AAAI'24)}                    & 50.4            & 52.6           & 60.9               & 68.4            & 62.1               & 63.3              & 41.6             & 43.1         & 42.5               & 49.7             & 51.5              & 55.4                   \\
\textbf{NPR (CVPR'24)}                     & 80.4             & 79.3           & 75.2             & 87.8            & 82.2            & 90.0           & 87.7                & 94.6              & 87.4              & 93.4            & \textbf{82.6}            & \textbf{89.0}            \\
\rowcolor[HTML]{C0C0C0} 
\textbf{LDR-Net (our)}                  & 83.7             & 85.1           & 85.5             & 94.2            & 87.1            & 94.4           & 83.5                & 92.9              & 91.1              & 97.9            & \textbf{\underline{86.2}}   & \textbf{\underline{92.9}}   \\ \hline
\end{tabular}
}
\vspace{-10pt}
\label{tab:cross_gan_evaluation_5}
\end{table*}

\subsection{Experiment Setup}
\textbf{Training Dataset:} In order to maintain consistency in our evaluation, we use the training set of ForenSynths \cite{wang2020cnn} to train our model. Based on previous research \cite{tan2024rethinking}, we select four different categories of this training set (cars, cats, chairs, and horses), each of which contains 18,000 synthetic images generated by ProGAN, as well as an equal number of real images selected from the LSUN \cite{yu2015lsun} dataset of an equal number of real images.

\textbf{Testing Dataset:} To evaluate the generalization ability of the proposed method in real-world scenarios, we use the following four testing datasets, which consists of various real images, diverse GAN and Diffusion models. 
\begin{itemize}
    \item \textbf{The ForenSynths dataset} \cite{wang2020cnn} includes images generated by eight models (ProGAN, StyleGAN, StyleGAN2, BigGAN, CycleGAN, StarGAN, GauGAN, and DeepFake) along with their corresponding real images.
    \item \textbf{The DiffusionForensics dataset} \cite{wang2023dire} contains images generated by eight diffusion models (ADM, DDPM, IDDPM, LDM, PNDM, VQ Diffusion, Stable Diffusion v1 (Sdv1), and Stable Diffusion v2 (Sdv2)), with real images sampled from the LSUN and ImageNet \cite{russakovsky2015imagenet} datasets.
    \item \textbf{The Ojha dataset} \cite{ojha2023towards} includes images generated by ADM, Glide, DALL-Emini, and LDM, with real images sourced from the LAION \cite{schuhmann2021400m} and ImageNet datasets.
    \item \textbf{The Self-Synthesized dataset} \cite{tan2024rethinking} contains images generated through 1000 diffusion steps (DDPM, IDDPM, and ADM) as well as synthetic content generated by Midjourney and DALLE \cite{ramesh2021zero}, collected from the social platform Discord.
\end{itemize}


\textbf{Baseline:} We compare our LDR-Net with SOTA methods, including NPR \cite{tan2024rethinking}, MI\_Net \cite{ba2024exposing}, CNNDetection \cite{wang2020cnn}, Frank \cite{frank2020leveraging}, Durall \cite{durall2020watch}, Patchfor \cite{chai2020makes}, F3Net \cite{qian2020thinking}, SelfBland \cite{shiohara2022detecting}, GANdetection \cite{mandelli2022detecting}, BiHPF \cite{jeong2022bihpf}, FrePGAN \cite{jeong2022frepgan}, LGrad \cite{tan2023learning}, and Ojha \cite{ojha2023towards}. In the experiments, we reimplement the baseline NPR \cite{tan2024rethinking} and MI\_Net \cite{ba2024exposing} using its official code, while other data are obtained from NPR \cite{tan2024rethinking}.

\textbf{Implemental Details:} We implement the proposed LDR-Net using the PyTorch framework and train it on an NVIDIA 3060Ti GPU.  The network is trained end-to-end using the Adam optimizer with binary cross-entropy as the loss function. The learning rate is set to 0.0002, the batch size is 32, and the model is trained for 40 epochs. We use two metrics for evaluation: average precision (AP) and accuracy (ACC).

\begin{table*}[]
\centering
\caption{The effect evaluation of Gaussian smoothing parameter $\sigma$ on model performance.}
\renewcommand\arraystretch{1}
\resizebox{\textwidth}{!}{ 

\begin{tabular}{ccccccccccccccccc|cc}
\hline
                                  & \multicolumn{2}{c}{\textbf{ProGAN}} & \multicolumn{2}{c}{\textbf{StyleGAN}}       & \multicolumn{2}{c}{\textbf{StyleGAN2}}      & \multicolumn{2}{c}{\textbf{BigGAN}}        & \multicolumn{2}{c}{\textbf{CycleGAN}} & \multicolumn{2}{c}{\textbf{StarGAN}}      & \multicolumn{2}{c}{\textbf{GauGAN}}   & \multicolumn{2}{c|}{\textbf{Deepfake}}      & \multicolumn{2}{c}{\textbf{Mean}} \\ \cline{2-19} 
\multirow{-2}{*}{\textbf{ $\sigma$}} & \textbf{ACC}      & \textbf{AP}     & \textbf{ACC}          & \textbf{AP}         & \textbf{ACC}          & \textbf{AP}         & \textbf{ACC}          & \textbf{AP}        & \textbf{ACC}       & \textbf{AP}      & \textbf{ACC}         & \textbf{AP}        & \textbf{ACC}       & \textbf{AP}      & \textbf{ACC}          & \textbf{AP}         & \textbf{ACC}    & \textbf{AP}     \\ \hline
\textbf{0.5}                      & 99.8              & 100.0           & 97.3                  & 99.8                & 95.1                  & 99.5                & 84.6                  & 90.2               & 87.8               & 94.8             & 96.1                 & 99.8               & 71.8               & 71.7             & 68.3                  & 69.1                & 87.6            & 90.6            \\

\textbf{1}             & 99.9         & 100.0                & 97.9              & 99.9                    & 96.5               & 99.8                   & 91.4             & 97.1                    & 91.4           & 97.6                 & 99.7                  & 100.0                & 80.8           & 86.7                 & 68.6          & 81.5                                              & \textbf{90.8}            & \textbf{95.3}            \\

\textbf{2}                        & 99.9              & 100.0           & 98.3                  & 99.9                & 95.5                  & 99.6                & 87.5                  & 93.4               & 83.7               & 92.3             & 98.4                 & 100.0              & 74.5               & 76.5             & 71.9                  & 75.1                & 88.7            & 92.1            \\ \hline
                      
\end{tabular}
}
\label{tab:effect}
\end{table*}

\begin{table*}[]
\centering
\caption{Robustness evaluation on the DiffusionForensics dataset.}
\renewcommand\arraystretch{1.1}
\resizebox{\textwidth}{!}{ 
\begin{tabular}{ccccccccccccccccccc|cc}
\hline
                                         &                                                        &                                               & \multicolumn{2}{c}{\textbf{ADM}}                            & \multicolumn{2}{c}{\textbf{DDPM}}                            & \multicolumn{2}{c}{\textbf{IDDPM}}                          & \multicolumn{2}{c}{\textbf{LDM}}                             & \multicolumn{2}{c}{\textbf{PNDM}}                            & \multicolumn{2}{c}{\textbf{VQ-Diffusion}}                    & \multicolumn{2}{c}{\textbf{Sdv1}}                           & \multicolumn{2}{c|}{\textbf{Sdv2}}                          & \multicolumn{2}{c}{\textbf{Mean}}                                             \\ \cline{4-21} 
\multirow{-2}{*}{\textbf{Manipulation}}  & \multirow{-2}{*}{\textbf{Parameter}}                   & \multirow{-2}{*}{\textbf{Method}}             & \textbf{ACC}                 & \textbf{AP}                  & \textbf{ACC}                 & \textbf{AP}                   & \textbf{ACC}                 & \textbf{AP}                  & \textbf{ACC}                 & \textbf{AP}                   & \textbf{ACC}                 & \textbf{AP}                   & \textbf{ACC}                 & \textbf{AP}                   & \textbf{ACC}                 & \textbf{AP}                  & \textbf{ACC}                 & \textbf{AP}                  & \textbf{ACC}                          & \textbf{AP}                           \\ \hline
                                         &                                                        & \textbf{NPR (CVPR'24)}                                  & 83.3                         & 93.9                         & 73.8                        &95.1                          & 64.4                         & 96.9                         & 99.3                        & 100.0                          & 87.2                         & 99.5                          & 99.8                         & 100.0                          & 83.7                         & 97.5                         & 99.4                         & 100.0                         & 86.4                                  & 97.9                                  \\
                                         &                                                        & \textbf{MI\_Net (AAAI'24)}                              & 71.5                         & 84.1                         & 62.8                         & 51.8                          & 50.5                         & 65.6                         & 94.7                         & 99.6                          & 52.6                         & 84.3                          & 99.4                         & 100.0                         & 71.2                         & 93.6                         & 98.4                         & 99.9                         & 75.1                                  & 84.9                                  \\
                                         & \multirow{-3}{*}{\textbf{7*7}}                         & \cellcolor[HTML]{C0C0C0}\textbf{LDR-Net (our)} & \cellcolor[HTML]{C0C0C0}86.9 & \cellcolor[HTML]{C0C0C0}95.2 & \cellcolor[HTML]{C0C0C0}88.2 & \cellcolor[HTML]{C0C0C0}98.2  & \cellcolor[HTML]{C0C0C0}83.3 & \cellcolor[HTML]{C0C0C0}98.7 & \cellcolor[HTML]{C0C0C0}99.7 & \cellcolor[HTML]{C0C0C0}100.0 & \cellcolor[HTML]{C0C0C0}88.9 & \cellcolor[HTML]{C0C0C0}99.4  & \cellcolor[HTML]{C0C0C0}99.7 & \cellcolor[HTML]{C0C0C0}100.0  & \cellcolor[HTML]{C0C0C0}89.7 & \cellcolor[HTML]{C0C0C0}98.1 & \cellcolor[HTML]{C0C0C0}99.6 & \cellcolor[HTML]{C0C0C0}100.0 & \cellcolor[HTML]{C0C0C0}\textbf{92.0} & \cellcolor[HTML]{C0C0C0}\textbf{98.7} \\ \cline{2-21} 
                                         &                               & \textbf{NPR (CVPR'24)}                                  & 81.0                         & 90.5                         & 83.3                         & 93.4                         & 80.8                         & 96.6                         & 98.0                         & 99.8                        & 83.8                         & 96.8                         & 97.3                        & 99.7                         & 79.6                         & 94.4                         & 98.0                         & 99.8                        & 87.7                                  & 96.4                         \\
                                         &                              & \textbf{MI\_Net (AAAI'24)}                              & 66.6                         & 81.0                         & 62.5                         & 43.2                          & 50.1                         & 58.0                         & 83.0                         & 99.1                          & 50.5                         & 73.8                          & 96.2                         & 99.9                          & 64.1                         & 92.8                         & 96.7                         & 99.9                         & 71.2                                  & 81.0                                  \\
\multirow{-6}{*}{\textbf{Gaussian Blur}} & \multirow{-3}{*}{\textbf{9*9}} & \cellcolor[HTML]{C0C0C0}\textbf{LDR-Net (our)} & \cellcolor[HTML]{C0C0C0}82.2 & \cellcolor[HTML]{C0C0C0}92.6 & \cellcolor[HTML]{C0C0C0}91.5 & \cellcolor[HTML]{C0C0C0}96.1 & \cellcolor[HTML]{C0C0C0}93.2 & \cellcolor[HTML]{C0C0C0}97.8 & \cellcolor[HTML]{C0C0C0}95.9 & \cellcolor[HTML]{C0C0C0}100.0 & \cellcolor[HTML]{C0C0C0}94.3 & \cellcolor[HTML]{C0C0C0}98.0 & \cellcolor[HTML]{C0C0C0}95.9 & \cellcolor[HTML]{C0C0C0}100.0 & \cellcolor[HTML]{C0C0C0}85.4 & \cellcolor[HTML]{C0C0C0}95.8 & \cellcolor[HTML]{C0C0C0}95.8 & \cellcolor[HTML]{C0C0C0}99.9 & \cellcolor[HTML]{C0C0C0}\textbf{91.8} & \cellcolor[HTML]{C0C0C0}\textbf{97.5}          \\ \hline
                                         &                                                        & \textbf{NPR (CVPR'24)}                                  & 80.7                         & 90.4                         & 81.4                         & 87.4                          & 66.5                         & 83.7                         & 97.9                         & 99.8                          & 82.2                         & 94.6                          & 98.2                         & 99.9                          & 87.9                         & 97.3                         & 98.0                         & 99.9                         & 86.6                                  & 94.1                                  \\
                                         &                                                        & \textbf{MI\_Net (AAAI'24)}                              & 59.2                         & 68.8                         & 61.8                         & 39.8                          & 50.1                         & 50.8                         & 77.8                         & 95.4                          & 50.7                         & 64.0                          & 78.2                         & 95.6                          & 56.5                         & 85.7                         & 81.3                         & 96.1                         & 64.5                                  & 74.5                                  \\
                                         & \multirow{-3}{*}{\textbf{0.5}}                         & \cellcolor[HTML]{C0C0C0}\textbf{LDR-Net (our)} & \cellcolor[HTML]{C0C0C0}87.4 & \cellcolor[HTML]{C0C0C0}95.7 & \cellcolor[HTML]{C0C0C0}96.9 & \cellcolor[HTML]{C0C0C0}99.5  & \cellcolor[HTML]{C0C0C0}88.8 & \cellcolor[HTML]{C0C0C0}98.9 & \cellcolor[HTML]{C0C0C0}99.4 & \cellcolor[HTML]{C0C0C0}100.0 & \cellcolor[HTML]{C0C0C0}93.8 & \cellcolor[HTML]{C0C0C0}99.4  & \cellcolor[HTML]{C0C0C0}98.7 & \cellcolor[HTML]{C0C0C0}99.9  & \cellcolor[HTML]{C0C0C0}86.6 & \cellcolor[HTML]{C0C0C0}97.6 & \cellcolor[HTML]{C0C0C0}99.0 & \cellcolor[HTML]{C0C0C0}99.9 & \cellcolor[HTML]{C0C0C0}\textbf{93.8} & \cellcolor[HTML]{C0C0C0}\textbf{98.9} \\ \cline{2-21} 
                                         &                               & \textbf{NPR (CVPR'24)*}                                  & 85.0                         & 98.9                         & 91.5                         & 100.0                         & 74.3                         & 99.2                         & 99.9                         & 100.0                         & 92.5                         & 100.0                         & 100.0                        & 100.0                         & 97.3                         & 99.9                         & 99.7                         & 100.0                        & 92.5                                  & \textbf{99.7}                         \\
                                         &                               & \textbf{MI\_Net (AAAI'24)*}                              & 75.8                         & 83.3                         & 61.5                         & 42.4                          & 49.2                         & 47.7                         & 96.7                         & 99.1                          & 60.1                         & 87.3                          & 97.0                         & 98.9                          & 84.2                         & 95.2                         & 97.8                         & 99.5                         & 77.8                                  & 81.7                                  \\
\multirow{-6}{*}{\textbf{Resizing}}         & \multirow{-3}{*}{\textbf{1.5}} & \cellcolor[HTML]{C0C0C0}\textbf{LDR-Net (our)} & \cellcolor[HTML]{C0C0C0}92.6 & \cellcolor[HTML]{C0C0C0}99.0 & \cellcolor[HTML]{C0C0C0}99.2 & \cellcolor[HTML]{C0C0C0}100.0 & \cellcolor[HTML]{C0C0C0}94.1 & \cellcolor[HTML]{C0C0C0}99.4 & \cellcolor[HTML]{C0C0C0}99.4 & \cellcolor[HTML]{C0C0C0}100.0 & \cellcolor[HTML]{C0C0C0}99.3 & \cellcolor[HTML]{C0C0C0}100.0 & \cellcolor[HTML]{C0C0C0}99.4 & \cellcolor[HTML]{C0C0C0}100.0 & \cellcolor[HTML]{C0C0C0}85.3 & \cellcolor[HTML]{C0C0C0}98.9 & \cellcolor[HTML]{C0C0C0}97.2 & \cellcolor[HTML]{C0C0C0}99.7 & \cellcolor[HTML]{C0C0C0}\textbf{95.8} & \cellcolor[HTML]{C0C0C0}99.6          \\ \hline
\end{tabular}
}
\label{robust}
\end{table*}

\subsection{Ablation Study}
To evaluate the contribution of each module, we use the LGA and LVP modules independently for generated image detection. We perform our experiments on the ForenSynths dataset. As shown in Table \ref{tab:ablation}, the complete LDR-Net achieves better performance compared to the standalone modules. Specifically, compared to LGA, LDR-Net improves the average ACC by 5.4\% and AP by 2.9\%. Similarly, compared to LVP, LDR-Net achieves an average improvement of 5.0\% in ACC and 5.4\% in AP across all datasets. More detailed results of the ablation experiments on the additional datasets are provided in the Supplementary Material.

This phenomenon can be attributed to the complementary nature of the LGA and LVP modules in capturing different aspects of generated image anomalies. Specifically, the LGA module focuses on detecting excessive smoothness or blurred textures in generated images by calculating the local gradient autocorrelation, which effectively highlights anomalies in edge and texture regions. However, relying solely on LGA fails to capture broader or more diverse forgery cues, such as pixel coding patterns. The LVP module captures the relative variation patterns between pixels through directional encoding, revealing the lack of natural complexity and diversity in generated images. Nevertheless, using only LVP fails to consider the prominent anomalies in high-frequency regions of generated images. The complete LDR-Net integrates both LGA and LVP features, capturing the differences between generated and real images from both edge-texture features and pixel distribution patterns. Consequently, the full LDR-Net achieves significantly higher detection accuracy than the standalone LGA or LVP modules.

To further evaluate the impact of different filters used in LGA module, we compare the performance of LDR-Net using three filters: Sobel, Roberts, and Canny. As shown in Table \ref{tab:ablation}, LDR-Net (Sobel) achieves the best detection performance compared to Roberts and Canny. The Sobel operator combines smoothing effects in gradient computation, effectively suppressing noise while preserving edge information, making it more sensitive to subtle edge and texture anomalies. In contrast, the Roberts filter uses a 2×2 kernel to compute diagonal gradients, which struggles to capture fine details. The Canny filter is sensitive to parameter settings, may discard weak but meaningful edges. These factors make Sobel more suitable for feature extraction in the LGA module.

\subsection{GAN-Sources Evaluation}
To validate the generalization capability of LDR-Net on GAN-generated images, we conduct evaluations on the ForenSynths dataset \cite{wang2020cnn}, which includes samples from eight different GAN-generative models. As shown in Table \ref{tab:cross_gan_evaluation_1}, our LDR-Net achieves state-of-the-art performance in terms of ACC, with an average ACC of 90.8\%. However, LDR-Net performs slightly lower than Ojha \cite{ojha2023towards} in terms of AP, achieving an average AP of 95.3\%, which is the second-best result among all methods. These results demonstrate that LDR-Net exhibits strong generalization across diverse GAN models. 

\subsection{Diffusion-Sources Evaluation}
To evaluate the performance of LDR-Net on unseen diffusion-generated images, we test it on the DIRE \cite{wang2023dire} and Ojha \cite{wang2023dire} datasets. From Table \ref{tab:cross_gan_evaluation_3}, we can find that LDR-Net is superior to SOTA methods, with an average ACC at least 2.2\% higher and an average AP at least 0.6\% higher than comparison methods. From Table \ref{tab:cross_gan_evaluation_4}, LDR-Net achieves an average ACC and AP that are at least 2.2\% and 0.5\% higher than SOTA methods. These findings highlight the superior adaptability and generalization of LDR-Net across unseen diffusion models.

To address the limitations of low-resolution and less realistic images generated with fewer diffusion steps in the Ojha dataset, we perform further evaluations on the Self-Synthesis dataset \cite{tan2024rethinking}, which includes diffusion images generated with 1000 steps. The results in Table \ref{tab:cross_gan_evaluation_5} show that LDR-Net achieves gains of 3.6\% and 3.9\% in average ACC and AP compared to NPR \cite{tan2024rethinking}. These results demonstrate that even when dealing with generated images through more extended diffusion processes, LDR-Net maintains exceptional generalization performance.

\subsection{Impact of Hyperparameter Selection}
To investigate the impact of the Gaussian smoothing standard deviation $\sigma$ on LDR-Net performance, we conduct experiments with $\sigma=0.5$, $\sigma=1$, and $\sigma=2$ on the ForenSynths dataset. The results in Table \ref{tab:effect} demonstrate that the model obtains the best detection performance when $\sigma=1$. Smaller value of $\sigma=0.5$ fails to suppress high-frequency noise in generated images, resulting in the lack of emphasis on high-frequency smoothing anomalies during the LGA computation process. Therefore, the sensitivity of LGA to anomalies in generated images will be reduced. Larger value of $\sigma=2$, while capable of suppressing noise, causes over-smoothing that removes critical features such as edges and textures in generated images, weakening the detection capability of the LGA module. In contrast, a moderate $\sigma=1$ achieves great balance between noise suppression and detail preservation, enabling the LGA module to capture more accurate features. 

\subsection{Robustness Evaluation}


\begin{figure}[tbp]
\centering
\includegraphics[width=1\columnwidth]{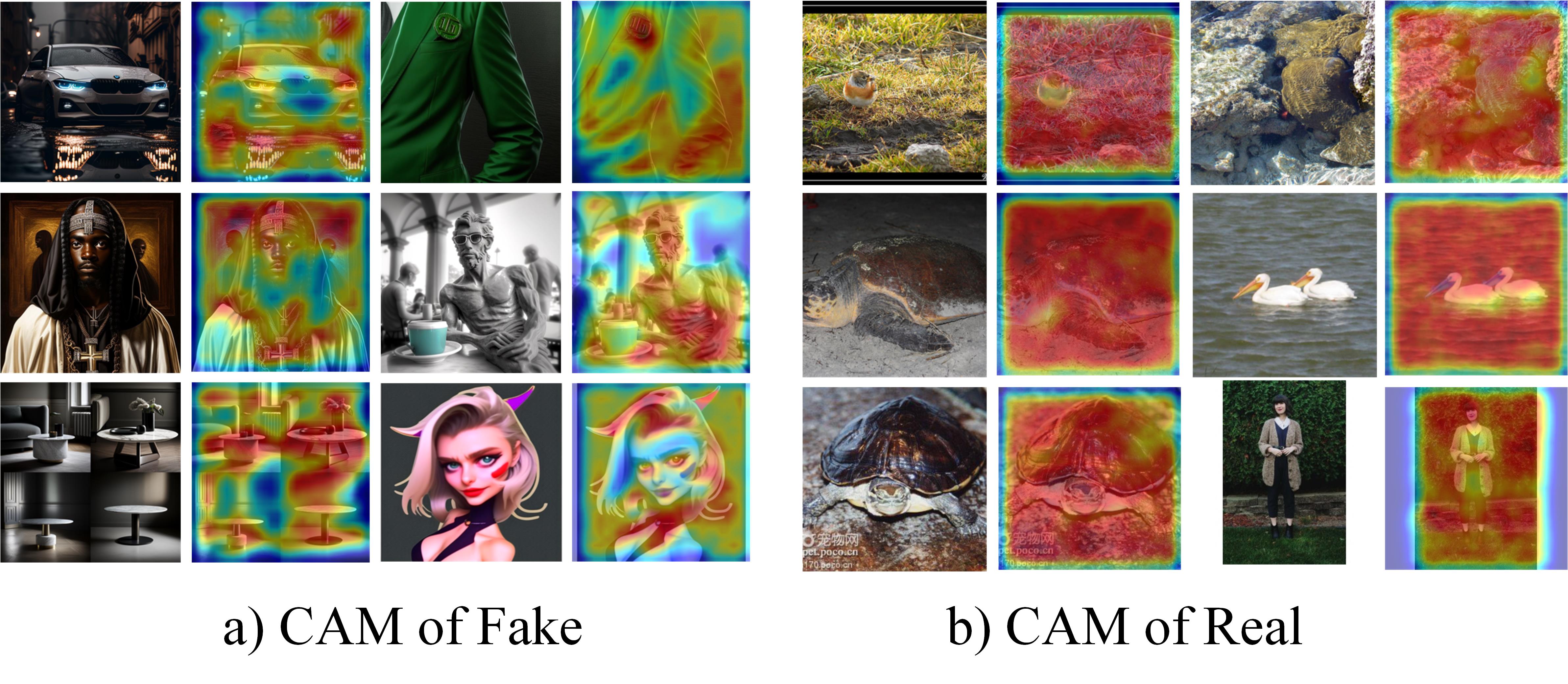}
\caption{CAM visualization for real and generated images.}
\label{visual1}
\end{figure}

To validate the robustness of LDR-Net, we evaluate its performance under Gaussian blur and image resizing using the DiffusionForensics dataset. The results of JPEG compression are provided in the Supplementary Material. As shown in Table \ref{robust}, for Gaussian blur (7×7 and 9×9 kernels), LDR-Net achieves an average ACC of 91.90\% and AP of 98.10\%, surpassing NPR by 4.85\% and 0.95\%, and MI\_Net by 18.75\% and 15.15\%, respectively. For resizing (scaling factors 0.5 and 1.5), LDR-Net achieves an average ACC of 94.80\% and an average AP of 99.25\%, exceeding NPR by 5.25\% and 2.35\%, and MI\_Net by 26.95\% and 21.50\%. These results demonstrate the superior adaptability of LDR-Net to post-processing operations.

This robustness stems from LDR-Net's emphasis on localized features and relative relationships, as opposed to absolute pixel values or global features, which are more vulnerable to the negative effects of post-processing. 


\subsection{Qualitative Analysis}
To further explore the inner characteristics of LDR-Net, we conduct a qualitative analysis using Class Activation Map (CAM) visualizations \cite{zhou2016learning}, with data sourced from Midjourney and DALLE. As presented in Fig. \ref{visual1}, the CAMs for real images highlights broader regions of the image, whereas the CAMs for AI-generated images focus on more local forged areas. These observations provide strong evidence of LDR-Net to effectively identify and localize generation artifacts, showcasing its robustness and precision in distinguishing between real and generated content.



\section{Conclusion}
This work proposes a novel localized discrepancy representation network (LDR-Net) to expose generation artifacts in AI-generated images. We employ advanced local feature extraction modules to enhance detection capabilities by capturing smoothing anomalies and unnatural pixel variation patterns. Our research indicates that integrating local gradient autocorrelation (LGA) with local variation pattern (LVP) provides a more comprehensive representation of directional gradient and pixel coding patterns differences between generated and real images. Experimental results demonstrate that LDR-Net exhibits strong generalization in detecting various generative models and unseen data distributions. Furthermore, LDR-Net illustrates robustness against post-processing operations.

\bibliographystyle{named}
\bibliography{ijcai25}

\end{document}